\documentclass[runningheads]{llncs}
\usepackage{graphicx}

\usepackage{tikz}
\usepackage{comment}
\usepackage{amsmath,amssymb} 
\usepackage{color}

\usepackage[accsupp]{axessibility}  

\usepackage{multirow}
\usepackage{enumitem}
\usepackage{romannum}
\usepackage{booktabs}
\usepackage{amsmath}

\usepackage{array}
\newcolumntype{P}[1]{>{\centering\arraybackslash}p{#1}}
\usepackage{hyperref}
\hypersetup{
	colorlinks = true, 
	urlcolor = magenta, 
	linkcolor = red, 
	citecolor = blue}
	
\usepackage{xcolor,colortbl}
\definecolor{Gray}{gray}{0.9}
\usepackage{cite}
\usepackage{wrapfig}

\begin{document}
\pagestyle{headings}
\mainmatter
\def\ECCVSubNumber{4697}  

\title{Tackling Background Distraction in\\Video Object Segmentation} 


\titlerunning{Tackling Background Distraction in Video Object Segmentation}
\author{Suhwan Cho$^1$\quad Heansung Lee$^1$\quad Minhyeok Lee$^1$\quad Chaewon Park$^1$\\Sungjun Jang$^1$\quad Minjung Kim$^1$\quad Sangyoun Lee$^{1,2}$}
\authorrunning{S. Cho et al.}
\institute{Yonsei University\and Korea Institute of Science and Technology (KIST)}
\maketitle

\begin{abstract}
Semi-supervised video object segmentation (VOS) aims to densely track certain designated objects in videos. One of the main challenges in this task is the existence of background distractors that appear similar to the target objects. We propose three novel strategies to suppress such distractors: 1) a spatio-temporally diversified template construction scheme to obtain generalized properties of the target objects; 2) a learnable distance-scoring function to exclude spatially-distant distractors by exploiting the temporal consistency between two consecutive frames; 3) swap-and-attach augmentation to force each object to have unique features by providing training samples containing entangled objects. On all public benchmark datasets, our model achieves a comparable performance to contemporary state-of-the-art approaches, even with real-time performance. Qualitative results also demonstrate the superiority of our approach over existing methods. We believe our approach will be widely used for future VOS research. Code and models are available at \url{https://github.com/suhwan-cho/TBD}.

\keywords{Video object segmentation, Metric learning, Temporal consistency, Video data augmentation}
\end{abstract}

\section{Introduction}
Video object segmentation (VOS) aims to keep tracking certain designated objects within an entire given video at the pixel level. Depending on the type of guidance provided with regard to the target objects, VOS can be divided into semi-supervised VOS, unsupervised VOS, referring VOS, and other subcategories. We focus on the semi-supervised setting, in which densely-annotated objects are provided in the initial frame of a video. Given its diverse applicability, semi-supervised VOS has recently attracted extensive attention in many vision fields such as autonomous driving, video editing, and video surveillance.

In semi-supervised VOS, one of the main challenges is the existence of background distractors that have a similar appearance to the target objects. As comparing visual properties is a fundamental technique to detect and track the designated objects, visual distractions can severely lower the reliability of a system. We propose three novel strategies to suppress the negative influence of background distractions in VOS: 1) a spatio-temporally diversified template construction scheme to prepare various object properties for reliable and stable prediction; 2) a learnable distance-scoring function to consider the temporal consistency of a video; 3) swap-and-attach data augmentation to provide hard training samples showing severe occlusions.

As VOS is a pixel-level classification task that requires fine-grained information, most feature matching-based approaches including VideoMatch~\cite{VideoMatch} and RANet~\cite{RANet} adopt pixel-level fine matching. While this approach is popular owing to its superior ability to capture details, it also has the shortcoming of easily causing noise. Given that each element in its template has small receptive fields, it is very susceptible to background distractors. To address this issue, we propose a novel way to construct a coarse matching template by compressing the fine matching template considering the probability of each pixel location. Unlike the fine matching template, each element in the coarse matching template covers large receptive fields that are dynamically defined based on past predictions. By employing fine matching and coarse matching simultaneously, more stable predictions can be obtained thanks to improved spatial diversity. In addition, to construct various templates exploiting multiple temporal properties in a video, we operate multiple templates that are independently built and updated based on their own strategies. The proposed spatio-temporally diversified template construction enables the model to capture local details while simultaneously obtaining a clear distinction between foreground and background as well as learning the ability to utilize various time-specific properties.

While visual information is important, exploiting temporal consistency between adjacent frames is also important for a robust VOS. To leverage the temporal consistency of a video, FEELVOS~\cite{FEELVOS} and CFBI~\cite{CFBI} apply a square window when transferring the reference frame information to the query frame. RMNet~\cite{RMNet} reduces the search area based on the coarsely predicted segmentation mask of the query frame. These hard windowing methods, which totally exclude the non-candidates, are effective for capturing the locality of a video, but as the size of a window is a discrete value, a human-tuning process is required for setting the hyper-parameters, which makes the solution less elegant and incomplete. To overcome this issue, we propose a learnable distance-scoring function that takes the distance between two pixels of consecutive frames as its input and simply outputs a spatial distance score from 0 to 1. When transferring the information from a reference frame pixel to a query frame pixel, it outputs a low score for distant pixels and vice versa.

We also propose a novel data augmentation technique for VOS, termed swap-and-attach augmentation. By simply swapping the objects between multiple sequences and attaching the swapped objects on the frames, realistic video data with severe occlusions can be simulated. As this creates training snippets containing multiple entangled objects, the model can learn strong feature representations to force each object to have unique features. Furthermore, it generalizes the model by dramatically increasing the amount of VOS training data, which is scarce as compared to other vision tasks.

We validate the effectiveness of our method by evaluating it on three public VOS benchmark datasets, i.e., DAVIS 2016~\cite{DAVIS2016}, DAVIS 2017~\cite{DAVIS2017}, and YouTube-VOS 2018~\cite{YTVOS} datasets. On all datasets, our method achieves a competitive performance compared to the current state-of-the-art methods while maintaining a real-time inference speed. In particular, in complex scenarios that have background distractors, our approach is proven to be much more effective than state-of-the-art solutions. Note that all our experiments are implemented on a single GeForce RTX 2080 Ti GPU, which makes our method much more accessible than other state-of-the-art methods, which require powerful hardware resources. We believe our proposed three novel strategies will be widely used for future VOS research.

Our main contributions can be summarized as follows:
\begin{itemize}[leftmargin=0.2in]
	\item We introduce a spatio-temporally diversified template construction mechanism to prepare diverse features for stable and robust feature matching. 
	
	\item We propose an end-to-end learnable distance-scoring function to fully exploit the temporal consistency between two consecutive frames in a video.
	
	\item We present swap-and-attach augmentation to learn a strong VOS model by ensuring the diversity of training data and simulating occluded samples. 
	
	\item Our proposed approach achieves a competitive performance on the DAVIS and YouTube-VOS datasets, while maintaining a real-time performance.
\end{itemize}

\section{Related Work}
\noindent\textbf{Feature similarity matching.} Contemporary VOS methods track and segment the designated objects based on the similarity of the embedded features. Given that semi-supervised VOS is a pixel-level classification task, pixel-level fine matching is adopted in most cases. PML~\cite{PML} learns a pixel-wise embedding using the nearest neighbor classifier. VideoMatch~\cite{VideoMatch} produces foreground and background similarity maps using a soft matching layer. Extending these works, FEELVOS~\cite{FEELVOS} and CFBI~\cite{CFBI} exploit global matching and local matching to obtain long-term and short-term appearance information. RANet~\cite{RANet} maximizes the use of similarity maps from feature similarity matching by ranking and selecting conformable feature maps as per their importance. To fully exploit the information extracted from all previous frames, STM~\cite{STM} adopts a memory network by defining the past frames with object masks as external memory and the current frame as the query. The query and the memory are densely matched in feature space, covering all the space–time pixel locations. Extended from STM, KMN~\cite{KMN} proposes memory-to-query matching to reduce background distractions when transferring the memorized information by applying a 2D Gaussian kernel. EGMN~\cite{EGMN} exploits an episodic memory network where frames are stored as nodes so that cross-frame correlations can be captured effectively. These memory-based methods achieve admirable performance but face a serious issue: the size of memory increases over time, and therefore, they are not efficient for long videos. To address this, GC~\cite{GC} stores and updates a key–value mapping instead of storing all the key and value features of the past frames. AFB-URR~\cite{AFB-URR} introduces an adaptive feature bank update scheme to efficiently discard obsolete features and absorb new features based on a pre-defined policy.

\noindent\textbf{Leveraging temporal consistency.} As a video shares some common properties between the consecutive frames, there are several VOS methods that leverage this temporal consistency of a video. FEELVOS and CFBI apply a fixed-sized square window for a local feature similarity matching. When transferring information from the previous adjacent frame to the query frame, each query frame pixel can only refer to the region within a square window centered on the spatial location of that pixel. Similarly, RMNet~\cite{RMNet} also reduces the search area by applying a square window. An optical flow model first coarsely predicts the segmentation mask of the query frame by warping the previous frame segmentation mask. Then, a square window is generated and applied based on the coarsely predicted query frame segmentation mask. These hard windowing methods are effective at leveraging the temporal consistency of a video, but as the size of a window is a discrete value, it needs a human-tuning process and is also not end-to-end learnable.

\begin{figure*}[t]
	\centering
	\includegraphics[width=1.0\linewidth]{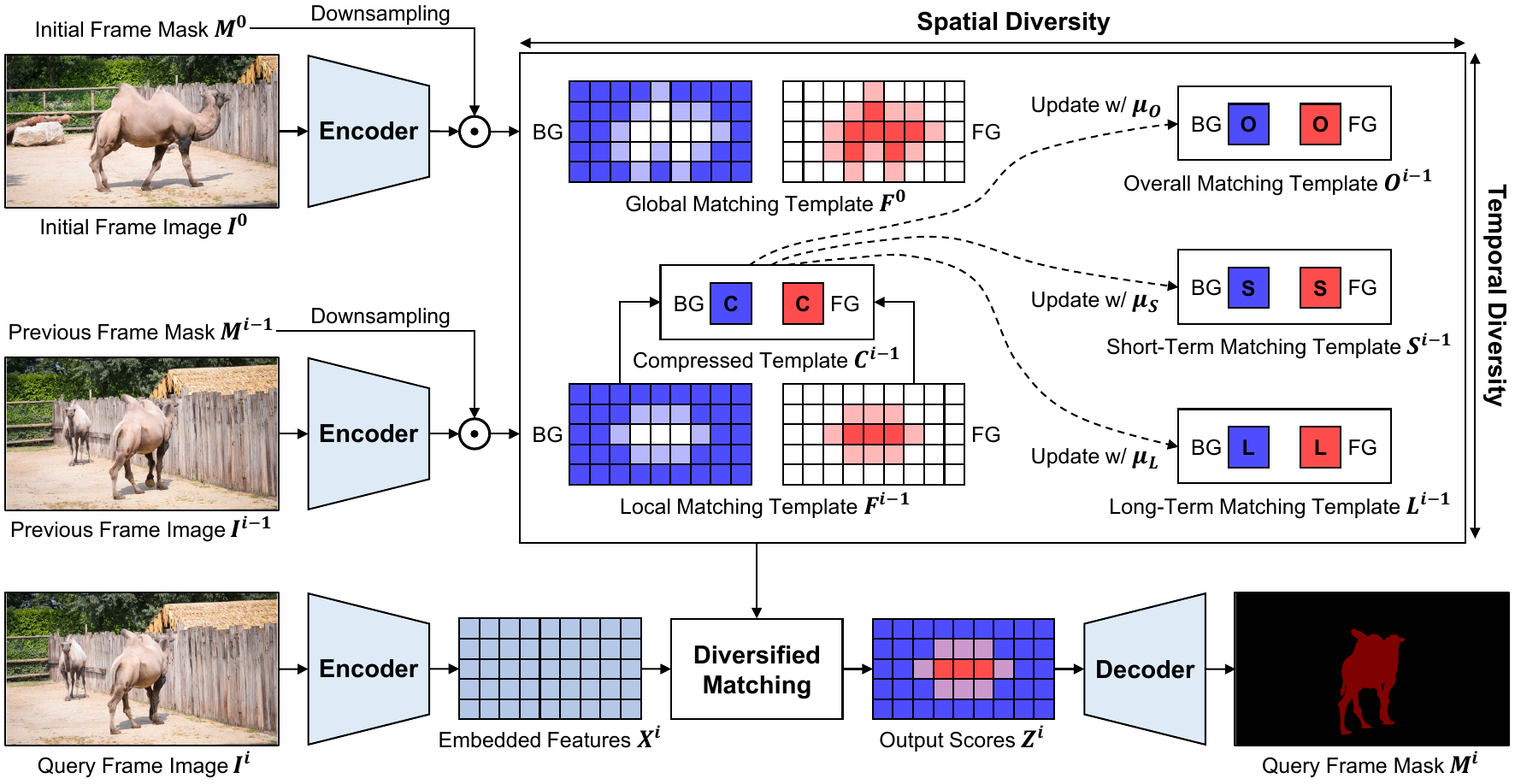}
	\caption{Architecture of our proposed method. The templates for feature similarity matching are independently constructed and updated as per their own objectives. The spatial diversity and temporal diversity of the templates are represented horizontally and vertically, respectively. Several components, e.g., skip connections and mask propagation, are omitted for better clarification of the overall architecture.}
	\label{figure1}
\end{figure*}

\section{Approach} 
In our framework, frames in a video sequence are segmented using the ground truth segmentation mask given in the initial frame. Mask prediction is performed based on feature similarity between the embedded features. To prepare various template features for feature matching, we employ a spatio-temporally diversified template construction mechanism. The results of feature matching are then fed into the decoder with low-level features extracted from the encoder and a downsampled previous adjacent frame mask. The overview of our framework is illustrated in Figure~\ref{figure1}.

\subsection{Deriving Spatial Diversity}
Following previous approaches~\cite{VideoMatch, FEELVOS, RANet, CFBI, BMVOS}, we employ pixel-level fine matching to capture the fine-grained information that is essential to generate detailed segmentations. Given that $I^i$ is an input image at frame $i$, the embedded features for that frame are defined as $X^i \in [-1,1]^{C \times H \times W}$ after channel L2 normalization. The provided or predicted segmentation mask and its downsampled version are denoted as $M^i \in [0, 1]^{2 \times H0 \times W0}$ and $m^i \in [0, 1]^{2 \times H \times W}$, respectively. The first channel indicates the probabilities of the background, and the second channel indicates those of a foreground object. Using $X^i$ and $m^i$, we define the pixel-level fine matching template $F^i \in [-1,1]^{C \times H \times W}$ as
\begin{align}
&F^i_{BG} = X^i \odot m^i_0\nonumber\\
&F^i_{FG} = X^i \odot m^i_1~,
\end{align}
where $\odot$ denotes the Hadamard product and $m^i_0$ and $m^i_1$ denote the first and second channels of $m^i$, respectively.

Fine matching is effective at capturing the details that are inherent at the pixel-level, but thereafter, is susceptible to background distractors that have a similar appearance to a foreground object. In order to obtain more general features for an object and its background, i.e., to obtain spatial diversity, we propose coarse matching to complement the fine matching. Unlike the fine matching template whose receptive fields are small and local, the receptive fields of the coarse matching template are dynamically defined covering a large range. Therefore, coarse matching is able to obtain a clear distinction between foreground and background because outlier features are less likely to negatively affect the feature similarity matching scores.

To design the coarse matching template, we first build a compressed template $C^i \in [-1,1]^{C \times 1 \times 1}$ that indicates the normalized mean features of $X^i$ weighted by the probabilities of each class in every spatial location. By using pre-computed $F^i$, it can be summarized as
\begin{align}
&C^i_{BG} = \mathcal{N}\left(\mathcal{S}\left(X^i \odot m^i_0\right)\right) = \mathcal{N}\left(\mathcal{S}\left(F^i_{BG}\right)\right)\nonumber\\
&C^i_{FG} = \mathcal{N}\left(\mathcal{S}\left(X^i \odot m^i_1\right)\right) = \mathcal{N}\left(\mathcal{S}\left(F^i_{FG}\right)\right)~,
\end{align}
where $\mathcal{S}$ and $\mathcal{N}$ indicate channel-wise summation and channel L2 normalization, respectively. The compressed template is used to build various coarse matching templates according to respective strategies that are pre-defined based on certain objectives to exploit various temporal characteristics.

\subsection{Deriving Temporal Diversity}
As much as spatial diversity is effective to enrich the semantic information of the embedded features, utilizing temporal diversity is also important considering the nature of a video. When constructing the fine matching templates, we use the initial frame to exploit accurate information and the previous adjacent frame to leverage the most related information, similar to FEELVOS~\cite{FEELVOS}, CFBI~\cite{CFBI}, and BMVOS~\cite{BMVOS}. Those matching templates are defined as the global matching template and local matching template, respectively.

As opposed to the fine matching template whose elements contain their unique positional information, the features in the compressed template do not contain any positional information because it only has a single spatial location. Therefore, it is able to integrate multiple compressed templates of historic frames based on specific goals. Using this, we build temporally diversified coarse matching templates to fully exploit various time-specific properties of a video. We define the coarse matching templates as follows: 1) an overall matching template that memorizes the general features along an entire video; 2) a short-term matching template that is built by paying more attention to temporally-near frames; 3) a long-term matching template that is built by paying more attention to temporally-distant frames. Assuming frame $i$ is the current frame being processed, the overall matching template $O^i \in [-1,1]^{C \times 1 \times 1}$, short-term matching template $S^i \in [-1,1]^{C \times 1 \times 1}$, and long-term matching template $L^i \in [-1,1]^{C \times 1 \times 1}$ are defined as follows.
\begin{eqnarray}
&\mu_{O,BG} = \Sigma^{i-1}_{k=0} \mathcal{S}\left(m^k_0\right) / \Sigma^{i}_{k=0} \mathcal{S}\left(m^k_0\right)\nonumber\\
&\mu_{O,FG} = \Sigma^{i-1}_{k=0} \mathcal{S}\left(m^k_1\right) / \Sigma^{i}_{k=0} \mathcal{S}\left(m^k_1\right)\nonumber\\[2pt]
&O^i_{BG} = \mathcal{N}\left(\mu_{O,BG} O^{i-1}_{BG} + \left(1 - \mu_{O,BG}\right) C^i_{BG}\right)\nonumber\\
&O^i_{FG} = \mathcal{N}\left(\mu_{O,FG} O^{i-1}_{FG} + \left(1 - \mu_{O,FG}\right) C^i_{FG}\right)\\[15pt]
&S^i_{BG} = \mathcal{N}\left(\mu_{S,BG} S^{i-1}_{BG} + \left(1 - \mu_{S,BG}\right) C^i_{BG}\right)\nonumber\\
&S^i_{FG} = \mathcal{N}\left(\mu_{S,FG} S^{i-1}_{FG} + \left(1 - \mu_{S,FG}\right) C^i_{FG}\right)\\[15pt]
&L^i_{BG} = \mathcal{N}\left(\mu_{L,BG} L^{i-1}_{BG} + \left(1 - \mu_{L,BG}\right) C^i_{BG}\right)\nonumber\\
&L^i_{FG} = \mathcal{N}\left(\mu_{L,FG} L^{i-1}_{FG} + \left(1 - \mu_{L,FG}\right) C^i_{FG}\right)
\end{eqnarray}

Considering that the compressed template is a channel normalized vector, taking the average over two vectors will generate a vector with a scale lower than 1, if two vectors have different directions. Therefore, the scale of the coarse matching templates will keep decreasing over time, which will lead to a scale-vanishing problem that engenders low confidence in the coarse matching. To prevent this, we apply channel re-normalization after every template update step to maintain the scale of the coarse matching templates at 1. The inertia values for short-term matching and long-term matching, i.e., $\mu_S$ and $\mu_L$, are learnable parameters and trained during the network training stage. To press the values to be trained for respective objectives, we set the initial values to $\frac{1}{1 + exp(1)}$ and $\frac{1}{1 + exp(-1)}$, respectively. At the initial frame, all $\mu$ values are set to 0 because there is no previous frame.

\subsection{Diversified Feature Similarity Matching}
Assuming frame $i$ is the query frame, the spatio-temporally diversified templates, i.e., global matching template $F^0$, local matching template $F^{i-1}$, overall matching template $O^{i-1}$, short-term matching template $S^{i-1}$, and long-term matching template $L^{i-1}$, are compared to the query frame embedded features $X^i$. Given that a feature similarity matching template has a size of $C\times M\times N$, taking a matrix inner product along the channel dimension with $X^i$ will output a similarity map with the size of $MN \times HW$. By taking query-wise maximum operation and reshaping, a score map with the size of $H \times W$ can be obtained for each template and class. As there are five different matching templates consisting of foreground and background, the output scores of feature similarity matching are defined as $Z^i \in [-1, 1]^{10 \times H \times W}$. These scores are fed into a decoder as an input, to provide visually derived information that can specify the target object.

\begin{figure*}[t]
	\centering
	\includegraphics[width=1.0\linewidth]{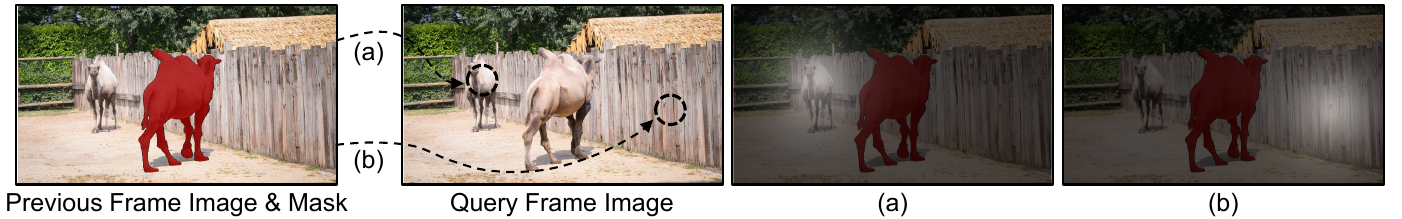}
	\caption{Visualized information transfer flow for local matching when spatial distance scoring is applied. The information of each reference frame pixel is weighted by a distance score that indicates how close two pixels are spatially.}
	\label{figure2}
\end{figure*}

\subsection{Spatial Distance Scoring}
As described before, the feature similarity matching operates in a non-local manner, which excludes the importance of temporal consistency of a video. However, adjacent frames in a video sequence are strongly dependent on each other, and therefore, are predominantly local. To exploit this locality of VOS, FEELVOS~\cite{FEELVOS} and CFBI~\cite{CFBI} restrict the search area by applying a square window for every spatial location when transferring the previous adjacent frame information to the query frame. Similarly, RMNet~\cite{RMNet} reduces the search area under a hard square that is generated by warping the previous frame mask using a pre-trained optical flow model. These hard windowing methods are effective at capturing the temporal consistency of a video, but as the size of a window is a discrete value, it needs a human-tuning process, which makes the solution less elegant and complete.

To obtain temporal consistency automatically without any human-tuning process, we propose an end-to-end learnable distance-scoring function that takes a spatial distance between two pixels as its input and simply outputs a distance score from 0 to 1. When performing local matching with the previous frame, we first calculate a distance matrix that comprises of the spatial distance of every previous adjacent frame pixel location $p$ and query frame pixel location $q$ based on Euclidean distance as
\begin{align}
&d = \sqrt{\left(p_x - q_x\right) ^ 2 + \left(p_y - q_y\right) ^ 2}~,
\end{align}
where $p_x$, $p_y$, $q_x$, and $q_y$ are $x$, $y$ coordinates of $p$ and $q$. Then, by independently applying the distance-scoring function to every single element in the matrix, spatial distance score $D$ can be obtained as
\begin{align}
&D = f\left(d; \rm{w}\right) = Sigmoid\left(\rm{w_2} * max\left(0, \rm{w_1} * d\right)\right)~,
\end{align}
where $f$ is the distance-scoring function and $\rm{w}$ indicates learnable parameters of $f$. By multiplying the spatial distance scores to feature similarity scores obtained from the local matching, the spatial distance as well as the feature distance can be reflected in the similarity matching scores. Figure~\ref{figure2} shows the information transfer flow when applying spatial distance scoring. As information transfer is adjusted by spatial distance scores between two pixels, temporal consistency between adjacent frames can be effectively considered.

\begin{figure*}[t]
	\centering
	\includegraphics[width=1.0\linewidth]{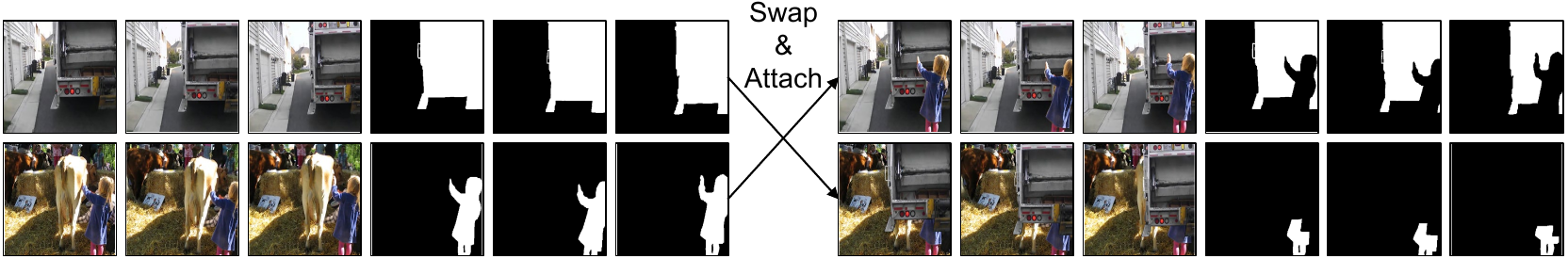}
	\caption{Example video sequences before and after applying swap-and-attach data augmentation method.}
	\label{figure3}
\end{figure*}

\subsection{Swap-and-Attach Augmentation}
As annotating video data in a pixel level is extremely laborious, VOS data is not sufficient compared to other computer vision tasks. To increase the amount of VOS training data, we propose a swap-and-attach data augmentation method. In Figure~\ref{figure3}, we show some example sequences when applying swap-and-attach augmentation. During the network training stage, each video sequence of a single batch exchanges its information with another video sequence of a different batch. The objects from multiple frames are jointly swapped between different sequences and attached on respective frames. As this process does not harm the temporal consistency of each video sequence, the generated video sequences are also realistic and have temporal connectivity between the consecutive frames. The swap-and-attach augmentation has two main advantages. First, as the amount of training data is dramatically increased, the model can be much more generalized. Second, as the objects may be severely occluded by the attached objects, the model can learn strong feature representations to force each object to have unique features.

\subsection{Implementation Details}
\noindent\textbf{Encoder.} We take the DenseNet-121~\cite{densenet} architecture as our encoder. To initialize the model with rich feature representations, we use the ImageNet~\cite{imagenet} pre-trained version. As the highest-level features extracted from original DenseNet-121 have low resolution, we skip the last block to use a higher feature resolution.

\noindent\textbf{Decoder.} Taking processed features from the encoder and cues from feature matching module, the decoder generates the final segmentation. To better exploit the locality of VOS, $m^{i-1}$ is fed into the decoder when predicting frame $i$. As $Z^i$ and $m^{i-1}$ have a small spatial size compared to the required output size, low-level features extracted from the encoder are also fed into the middle of the decoder via skip connections. For fast decoding, a channel-reducing deconvolutional layer~\cite{deconv} is added after every skip connection as in CRVOS~\cite{CRVOS}. To obtain rich feature representations, we adopt CBAM~\cite{CBAM} after every deconvolutional layer.

\subsection{Network Training}
Following most state-of-the-art methods~\cite{STM, KMN, AFB-URR, RMNet, LCM}, we simulate training samples using static image dataset, COCO~\cite{COCO}, to obtain sufficient training data. After pre-training the network using static images, either the DAVIS 2017~\cite{DAVIS2017} training set or YouTube-VOS 2018~\cite{YTVOS} training set is used for the main training depending on the testing dataset. From a video sequence, we randomly crop 384$\times$384 patches from 10 consecutive frames jointly. If the first frame does not contain any object, the cropping is repeated until it contains at least one object. After that, a single designated object is defined as a foreground object and other objects are considered as background. Similar to CFBI~\cite{CFBI}, if the number of foreground pixels is not enough, that sequence is not used for network training to prevent the model from being biased to background attributes. We use cross-entropy loss and Adam optimizer~\cite{adam} with a learning rate of 1e-4 without learning rate decay. During the network training, the backbone encoder is frozen and all batch normalization layers~\cite{batchnorm} are disabled. The swap-and-attach augmentation is applied with a probability of 20\%.

\section{Experiments}
In this section, we first describe the datasets and evaluation metrics used in this study in Section~\ref{setup}. Each component of our proposed approach is validated in Section~\ref{analysis}. Quantitative and qualitative comparison with state-of-the-art methods can be found in Section~\ref{quantitative} and Section~\ref{qualitative}, respectively. Our method is abbreviated as TBD.

\subsection{Experimental Setup}
\label{setup}
\noindent\textbf{Datasets.} The task of semi-supervised VOS is associated with two popular benchmark datasets, i.e., DAVIS~\cite{DAVIS2016, DAVIS2017} and YouTube-VOS~\cite{YTVOS} datasets. DAVIS 2016 consists of 30 videos in the training set and 20 videos in the validation set. DAVIS 2017 comprises 60 videos in the training set, 30 videos in the validation set, and 30 videos in the test-dev set. Each video is 24 fps and all frames are annotated. YouTube-VOS 2018 is the largest dataset for VOS, containing 3,471 videos in the training set and 474 videos in the validation set. All videos are 30 fps, annotated every five frames.

\noindent\textbf{Evaluation metrics.} We use the standard evaluation metrics for VOS. The predicted segmentation masks with hard labels are evaluated using region accuracy and contour accuracy. Region accuracy $\mathcal{J}$ can be obtained by computing the number of pixels in the intersection between the predicted segmentation mask and the ground truth segmentation mask and then dividing it by the size of their union. Contour accuracy $\mathcal{F}$ can be obtained using the same process but is evaluated only on the object boundaries. The overall accuracy $\mathcal{G}$ is the average of the $\mathcal{J}$ and $\mathcal{F}$ measures.

\begin{table}[t]
	\centering 
	\caption{Ablation study on the proposed components. G, L, O, and T indicate the use of global, local, overall, and short-term \& long-term matching templates, respectively. DS, HW, and Size denote spatial distance scoring, hard windowing, and window size for the hard windowing (the distance between the center point and side of a window). Aug indicates the use of swap-and-attach augmentation. The models are tested on the DAVIS 2017 validation set.}
	\begin{tabular}{P{1.6cm}|P{0.6cm}P{0.6cm}P{0.6cm}P{0.6cm}|P{0.8cm}P{0.8cm}P{0.8cm}|P{0.8cm}|P{0.8cm}P{0.8cm}P{0.8cm}}
		\toprule
		Version &G & L &O &T &DS &HW &Size &Aug &$\mathcal{G}_\mathcal{M}$ &$\mathcal{J}_\mathcal{M}$ &$\mathcal{F}_\mathcal{M}$\\
        \midrule
		\Romannum{1} &\checkmark &\checkmark & & &\checkmark & &- &\checkmark &77.2 &75.3 &79.1\\
		\Romannum{2} & & &\checkmark &\checkmark &\checkmark & &- &\checkmark &72.7 &70.2 &75.2\\
		\Romannum{3} &\checkmark &\checkmark &\checkmark & &\checkmark & &- &\checkmark &79.1 &76.7 &81.6\\
		\Romannum{4} &\checkmark &\checkmark & &\checkmark &\checkmark & &- &\checkmark &78.7 &76.7 &80.7\\
        \midrule
        \Romannum{5} &\checkmark &\checkmark &\checkmark &\checkmark & & &- &\checkmark &78.8 &76.7 &80.9\\
        \Romannum{6} &\checkmark &\checkmark &\checkmark &\checkmark & &\checkmark &1 &\checkmark &79.2 &76.9 &81.5\\
        \Romannum{7} &\checkmark &\checkmark &\checkmark &\checkmark & &\checkmark &2 &\checkmark &79.3 &76.9 &81.8\\
        \Romannum{8} &\checkmark &\checkmark &\checkmark &\checkmark & &\checkmark &4 &\checkmark &79.9 &77.4 &82.5\\
        \Romannum{9} &\checkmark &\checkmark &\checkmark &\checkmark & &\checkmark &8 &\checkmark &78.5 &76.3 &80.7\\
        \midrule
        \Romannum{10} &\checkmark &\checkmark &\checkmark &\checkmark &\checkmark & &- & &78.6 &76.1 &81.2\\
        \midrule
		\Romannum{11} &\checkmark &\checkmark &\checkmark &\checkmark &\checkmark & &- &\checkmark &80.0 &77.6 &82.3\\
		\bottomrule
	\end{tabular}
	\label{Table:ablation}
\end{table}

\subsection{Analysis}
\label{analysis}
\begin{figure*}[t]
	\centering
	\includegraphics[width=1.0\linewidth]{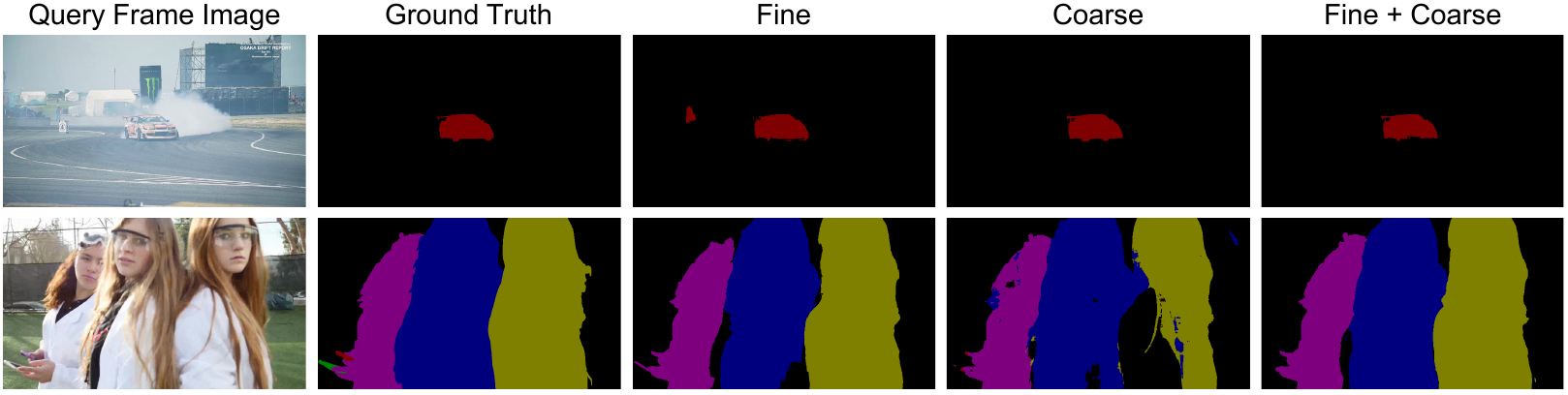}
	\caption{Qualitative comparison of various model versions with different spatial template construction schemes. Fine and coarse indicate the use of fine matching and coarse matching, respectively.}
	\label{figure4}
\end{figure*}

\noindent\textbf{Spatio-temporal diversity of template.} To capture the local keypoints while perceiving the global tendencies, we employ a fine matching template and coarse matching template simultaneously. In addition, we build each template using multiple sub-templates to exploit various temporal properties. In Table~\ref{Table:ablation}, various model versions with different template construction schemes are compared. As can be seen, a joint use of fine matching and coarse matching significantly improves the performance compared to when each one is used individually. The use of diverse temporal properties is also quantitatively validated to be effective. In Figure~\ref{figure4}, we compare the results of various model versions with different spatial diversities of a feature matching template. As can be seen, fine matching is effective for capturing local details (second sequence), but thereafter, is susceptible to visual distractions as it misses the global tendencies (first sequence). By contrast, coarse matching can capture the general features for an object (first sequence), but is not able to capture the details of the objects (second sequence). By using fine and coarse matching methods simultaneously, their respective advantages can be effectively leveraged.

\setlength\intextsep{0pt}
\begin{wrapfigure}{r}{0.5\textwidth}
\centering
\includegraphics[width=0.49\textwidth]{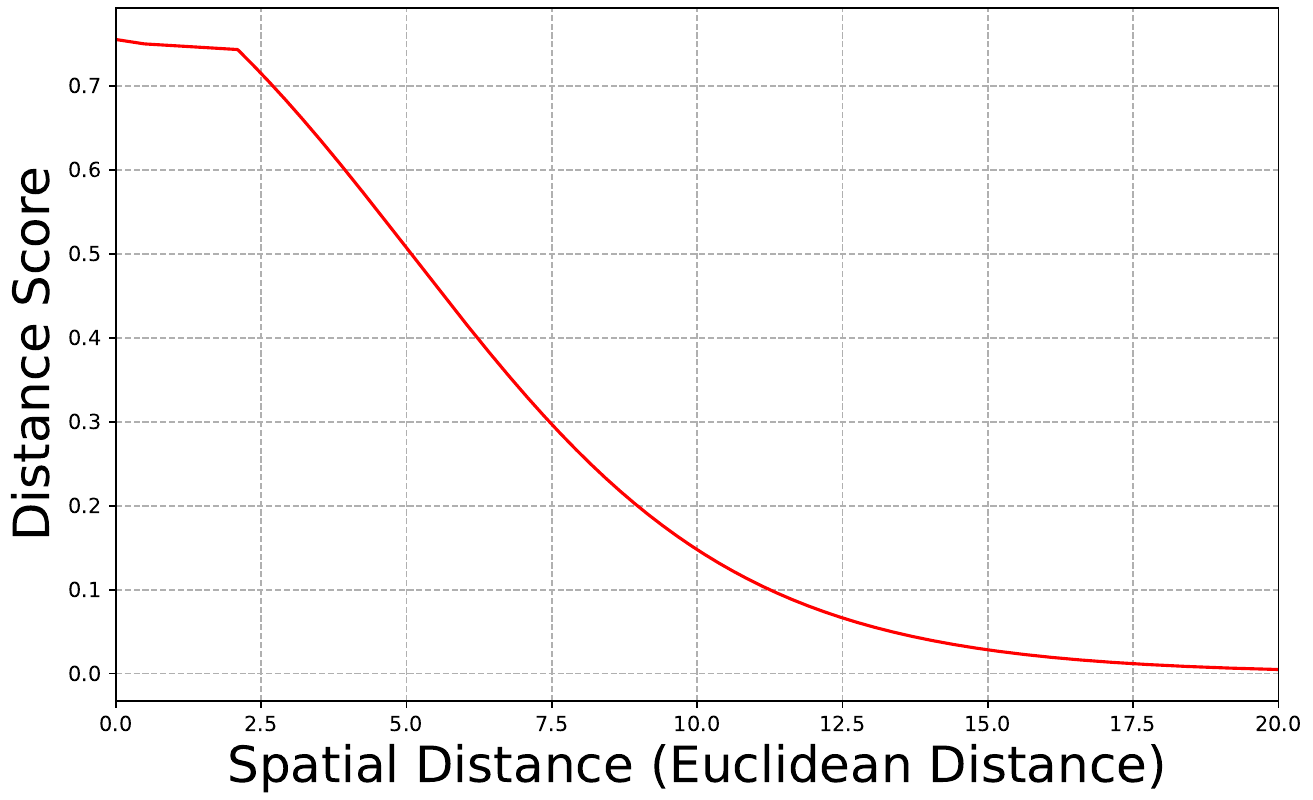}
\caption{Visualized input and output of a fully-trained distance-scoring function. Spatial distance indicates the Euclidean distance between two pixels after feature embedding.}
\label{figure5}
\end{wrapfigure}

\noindent\textbf{Spatial distance scoring.} In order to capture the temporal consistency between consecutive frames in a video without a human-tuning process, we present a novel spatial distance scoring method by learning a distance-scoring function that takes a spatial distance between two pixels as its input and outputs a naive distance score. According to Table~\ref{Table:ablation}, our proposed spatial distance scoring method is more effective than hard windowing method used in FEELVOS~\cite{FEELVOS} and CFBI~\cite{CFBI}, even without the need for manual tuning. We also visualize a fully-trained distance-scoring function to confirm that it really works. As can be seen in Figure~\ref{figure5}, the fully-trained distance-scoring function outputs a high distance score when two pixels are spatially near, and vice versa, which accords with the expected behavior.

\noindent\textbf{Swap-and-attach augmentation.} Our proposed swap-and-attach augmentation is also demonstrated to be effective through a quantitative evaluation in Table~\ref{Table:ablation}. Simply swapping and attaching the objects of different batches brings 1.4\% improvement on a $\mathcal{G}$ score, as it increases the amount of training data and provides hard training samples that contain some severe occlusions.

\begin{table}[t]
	\centering 
	\caption{Quantitative evaluation on the DAVIS datasets. OL denotes online learning. (+S) denotes the use of static image datasets during the network training.}
	\begin{tabular}{p{3cm}P{0.45cm}P{0.75cm}P{0.75cm}P{0.75cm}P{0.75cm}P{0.75cm}P{0.75cm}P{0.75cm}P{0.75cm}P{0.75cm}P{0.75cm}}
		\toprule
		\multicolumn{3}{c}{} &\multicolumn{3}{c}{2016 val} &\multicolumn{3}{c}{2017 val} &\multicolumn{3}{c}{2017 test-dev}\\
		\cline{4-12}
		Method &OL &fps &$\mathcal{G}_\mathcal{M}$ &$\mathcal{J}_\mathcal{M}$ &$\mathcal{F}_\mathcal{M}$ &$\mathcal{G}_\mathcal{M}$ &$\mathcal{J}_\mathcal{M}$ &$\mathcal{F}_\mathcal{M}$ &$\mathcal{G}_\mathcal{M}$ &$\mathcal{J}_\mathcal{M}$ &$\mathcal{F}_\mathcal{M}$\\
		\midrule
		STCNN~(+S)~\cite{STCNN} & &0.26 &83.8 &83.8 &83.8 &61.7 &58.7 &64.6 &- &- &-\\
		FEELVOS~(+S)~\cite{FEELVOS} &  &2.22 &81.7 &80.3 &83.1 &69.1 &65.9 &72.3 &54.4 &51.2 &57.5\\
		DMM-Net~(+S)~\cite{DMM-Net} & &- &- &- &- &70.7 &68.1 &73.3 &- &- &-\\
		AGSS-VOS~(+S)~\cite{AGSS-VOS} & &10.0 &- &- &- &67.4 &64.9 &69.9 &57.2 &54.8 &59.7\\
		RANet~(+S)~\cite{RANet} & &30.3 &85.5 &85.5 &85.4 &65.7 &63.2 &68.2 &55.3 &53.4 &57.2\\
		DTN~(+S)~\cite{DTN} & &14.3 &83.6 &83.7 &83.5 &67.4 &64.2 &70.6 &- &- &-\\
		STM~(+S)~\cite{STM} & &6.25 &86.5 &84.8 &\underline{88.1} &71.6 &69.2 &74.0 &- &- &-\\
		DIPNet~(+S)~\cite{DIPNet} & &0.92 &86.1 &85.8 &86.4 &68.5 &65.3 &71.6 &55.2 &- &-\\
        LWL~(+S)~\cite{LWL} &\checkmark &14.0 &- &- &- &74.3 &72.2 &76.3 &- &- &-\\
		CFBI~(+S)~\cite{CFBI} & &5.56 &86.1 &85.3 &86.9 &74.9 &72.1 &77.7 &- &- &-\\
		GC~(+S)~\cite{GC} & &25.0 &86.6 &\underline{87.6} &85.7 &71.4 &69.3 &73.5 &- &- &-\\
		KMN~(+S)~\cite{KMN} & &8.33 &\underline{87.6} &87.1 &\underline{88.1} &76.0 &74.2 &77.8 &- &- &-\\
        AFB-URR~(+S)~\cite{AFB-URR} & &4.00 &- &- &- &74.6 &73.0 &76.1 &- &- &-\\
        STG-Net~(+S)~\cite{STG-Net} & &- &85.7 &85.4 &86.0 &74.7 &71.5 &77.9 &\underline{63.1} &\underline{59.7} &\underline{66.5}\\
		RMNet~(+S)~\cite{RMNet} & &11.9 &81.5 &80.6 &82.3 &75.0 &72.8 &77.2 &- &- &-\\
        LCM~(+S)~\cite{LCM} & &8.47 &- &- &- &75.2 &73.1 &77.2 &- &- &-\\
        SSTVOS~(+S)~\cite{SSTVOS} & &- &- &- &- &78.4 &75.4 &81.4 &- &- &-\\
        HMMN~(+S)~\cite{HMMN} & &10.0 &\textbf{89.4} &\textbf{88.2} &\textbf{90.6} &\textbf{80.4} &\textbf{77.7} &\textbf{83.1} &- &- &-\\
		\rowcolor{Gray}
		TBD~(+S) & &\textbf{50.1} &86.8 &87.5 &86.2 &\underline{80.0} &\underline{77.6} &\underline{82.3} &\textbf{69.4} &\textbf{66.6} &\textbf{72.2}\\
		\hline
		AGSS-VOS~\cite{AGSS-VOS} & &10.0 &- &- &- &66.6 &63.4 &69.8 &54.3 &51.5 &57.1\\
		RANet~\cite{RANet} & &30.3 &- &73.2 &- &- &- &- &- &- &-\\
		STM~\cite{STM} & &6.25 &- &- &- &43.0 &38.1 &47.9 &- &- &-\\
		FRTM~\cite{FRTM} &\checkmark &21.9 &81.7 &- &- &68.8 &66.4 &71.2 &- &- &-\\
		TVOS~\cite{TVOS} & &37.0 &- &- &- &72.3 &69.9 &74.7 &\underline{63.1} &58.8 &\underline{67.4}\\
        JOINT~\cite{JOINT} &\checkmark &4.00 &- &- &- &\textbf{78.6} &\textbf{76.0} &\textbf{81.2} &- &- &-\\
		BMVOS~\cite{BMVOS} & &\underline{45.9} &\underline{82.2} &\underline{82.9} &\underline{81.4} &72.7 &70.7 &74.7 &62.7 &\underline{60.7} &64.7\\
		\rowcolor{Gray}
		TBD & &\textbf{50.1} &\textbf{84.3} &\textbf{85.2} &\textbf{83.4} &\underline{75.2} &\underline{73.2} &\underline{77.2} &\textbf{66.0} &\textbf{63.1} &\textbf{68.9}\\
		\hline
	\end{tabular}
	\label{Table:DAVIS}
\end{table}

\begin{table}[h!]
	\centering 
	\caption{Quantitative evaluation on the YouTube-VOS 2018 validation set. OL denotes online learning and (+S) denotes the use of static image datasets during the network training. Gen. denotes the generalization gap between seen and unseen categories.}
	\begin{tabular}{lP{0.8cm}P{1cm}P{1cm}P{1cm}P{1cm}P{1cm}P{1cm}P{1cm}}
		\toprule
		Method &OL &fps &$\mathcal{G}_\mathcal{M}$ &$\mathcal{J}_\mathcal{S}$ &$\mathcal{J}_\mathcal{U}$ &$\mathcal{F}_\mathcal{S}$ &$\mathcal{F}_\mathcal{U}$ &Gen.\\
		\midrule
        DMM-Net~(+S)~\cite{DMM-Net} & &12.0 &51.7 &58.3 &41.6 &60.7 &46.3 &15.6\\
		STM~(+S)~\cite{STM} & &- &79.4 &79.7 &72.8 &84.2 &80.9 &5.10\\
		SAT~(+S)~\cite{SAT} & &39.0 &63.6 &67.1 &55.3 &70.2 &61.7 &10.2\\
		LWL~(+S)~\cite{LWL} &\checkmark &- &81.5 &80.4 &76.4 &84.9 &84.4 &2.25\\
        EGMN~(+S)~\cite{EGMN} & &- &80.2 &80.7 &74.0 &85.1 &80.9 &5.45\\
		CFBI~(+S)~\cite{CFBI} & &- &81.4 &81.1 &75.3 &85.8 &83.4 &4.10\\
		GC~(+S)~\cite{GC} & &- &73.2 &72.6 &68.9 &75.6 &75.7 &\underline{1.80}\\
		KMN~(+S)~\cite{KMN} & &- &81.4 &81.4 &75.3 &85.6 &83.3 &4.20\\
        AFB-URR~(+S)~\cite{AFB-URR} & &- &79.6 &78.8 &74.1 &83.1 &82.6 &2.60\\
		STG-Net~(+S)~\cite{STG-Net} & &6.00 &73.0 &72.7 &69.1 &75.2 &74.9 &1.95\\
		RMNet~(+S)~\cite{RMNet} & &- &81.5 &82.1 &75.7 &85.7 &82.4 &4.85\\
		LCM~(+S)~\cite{LCM} & &- &82.0 &\underline{82.2} &75.7 &86.7 &83.4 &4.90\\
		GIEL~(+S)~\cite{GIEL} & &- &80.6 &80.7 &75.0 &85.0 &81.9 &4.40\\
		SwiftNet~(+S)~\cite{SwiftNet} & &- &77.8 &77.8 &72.3 &81.8 &79.5 &3.90\\
		SSTVOS~(+S)~\cite{SSTVOS} & &- &81.7 &81.2 &76.0 &- &- &-\\
		JOINT~(+S)~\cite{JOINT} &\checkmark &- &\underline{83.1} &81.5 &\textbf{78.7} &85.9 &\textbf{86.5} &\textbf{1.10}\\
		HMMN~(+S)~\cite{HMMN} & &- &82.6 &82.1 &76.8 &\underline{87.0} &84.6 &3.85\\
		AOT-T~(+S)~\cite{AOT} & &\underline{41.0} &80.2 &80.1 &74.0 &84.5 &82.2 &4.20\\
		SwinB-AOT-L~(+S)~\cite{AOT} & &9.30 &\textbf{84.5} &\textbf{84.3} &\underline{77.9} &\textbf{89.3} &\underline{86.4} &4.65\\
		STCN~(+S)~\cite{STCN} & &- &83.0 &81.9 &\underline{77.9} &86.5 &85.7 &2.40\\
		\rowcolor{Gray}
		TBD~(+S) & &30.4 &80.5 &79.4 &75.5 &83.8 &83.2 &2.25\\
		\hline
		RVOS~\cite{RVOS} & &22.7 &56.8 &63.6 &45.5 &67.2 &51.0 &17.2\\
        A-GAME~\cite{A-GAME} & &- &66.1 &67.8 &60.8 &- &- &-\\
		AGSS-VOS~\cite{AGSS-VOS} & &12.5 &71.3 &71.3 &65.5 &76.2 &73.1 &4.45\\
		CapsuleVOS~\cite{CapsuleVOS} & &13.5 &62.3 &67.3 &53.7 &68.1 &59.9 &10.9\\
		STM~\cite{STM} & &- &68.2 &- &- &- &- &-\\
		FRTM~\cite{FRTM} &\checkmark &- &72.1 &72.3 &65.9 &76.2 &74.1 &4.25\\
		TVOS~\cite{TVOS} & &37.0 &67.8 &67.1 &63.0 &69.4 &71.6 &\textbf{0.95}\\
		STM-cycle~\cite{STM-cycle} & &\textbf{43.0} &69.9 &71.7 &61.4 &75.8 &70.4 &7.85\\
		BMVOS~\cite{BMVOS} & &28.0 &\underline{73.9} &\underline{73.5} &\underline{68.5} &\underline{77.4} &\underline{76.0} &3.20\\
		\rowcolor{Gray}
		TBD & &30.4 &\textbf{77.8} &\textbf{77.4} &\textbf{72.7} &\textbf{81.2} &\textbf{79.9} &\underline{3.00}\\
		\hline
	\end{tabular}
	\label{Table:YTVOS}
\end{table}

\subsection{Quantitative Results}
\label{quantitative}
\noindent\textbf{DAVIS.} We present a quantitative comparison of our proposed method with state-of-the art methods on the DAVIS 2016~\cite{DAVIS2016} and DAVIS 2017~\cite{DAVIS2017} datasets in Table~\ref{Table:DAVIS}. For all methods, inference speed is calculated on the DAVIS 2016 validation set with the 480p resolution. TBD shows competitive performance on all datasets, even maintaining a real-time inference speed. Compared to the real-time methods faster than 24 fps (DAVIS default setting), it outperforms all existing methods by a significant margin.

\noindent\textbf{YouTube-VOS.} In Table~\ref{Table:YTVOS}, our method is compared to state-of-the-art methods on the YouTube-VOS 2018 validation set. Gen. indicates the generalization gap proposed in MAST~\cite{MAST}. In order to capture small objects, we use the 720p resolution if a target object contains less than 1,000 pixels; if not, we use the 480p resolution. The experimental results indicate TBD is comparable to other competitive methods, while showing a fast inference speed of 30.4 fps.

\begin{figure*}[t]
	\centering
	\includegraphics[width=1.0\linewidth]{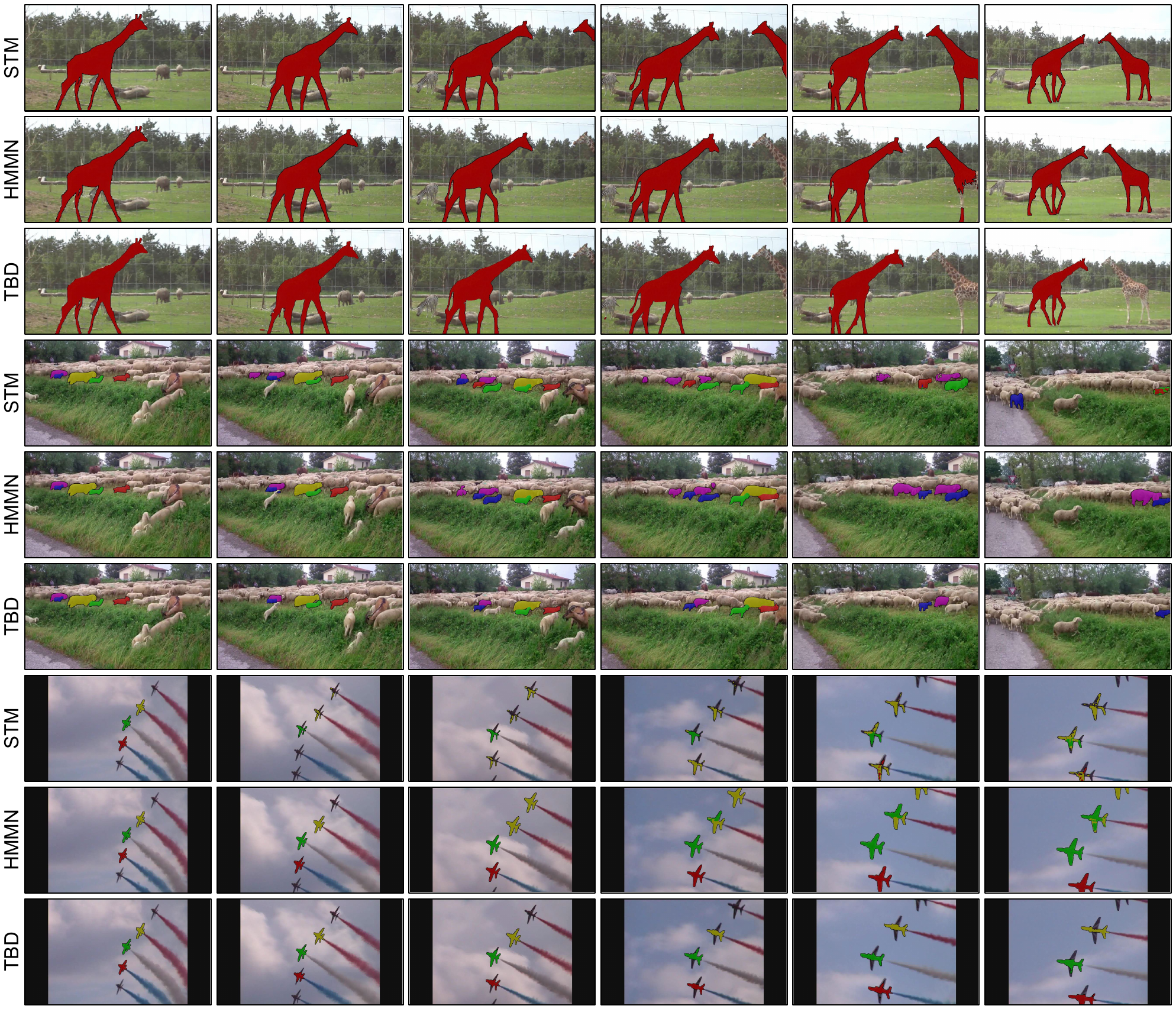}
	\caption{Qualitative evaluation of TBD obtained by comparing it to other state-of-the-art methods.}
	\label{figure6}
\end{figure*}

\subsection{Qualitative Results}
\label{qualitative}
In Figure~\ref{figure6}, we qualitatively compare our method to state-of-the-art STM~\cite{STM} and HMMN~\cite{HMMN}. As can be seen, TBD predicts more stable outputs compared to other methods, thanks to its diverse feature similarity matching templates and spatial distance scoring. Exceptional temporal consistency and robustness against occlusion can also be observed.

\section{Conclusion}
We introduce a spatio-temporally diversified template construction scheme for robust and stable VOS. In addition, a spatial distance scoring method with a learnable distance-scoring function and a novel swap-and-attach augmentation method are proposed. On public benchmark datasets, our approach achieves a comparable performance to the current state-of-the-art solutions while maintaining a real-time inference speed. We believe our approach will be widely used for future VOS research and various fields in computer vision.\\

\noindent\textbf{Acknowledgement.} This research was supported by R\&D program for Advanced Integrated-intelligence for Identification (AIID) through the National Research Foundation of KOREA (NRF) funded by Ministry of Science and ICT (NRF-2018M3E3A1057289) and the KIST Institutional Program (Project No.2E31051-21-203).

\clearpage
%
%
\bibliographystyle{splncs04}
\bibliography{egbib}
\end{document}